\documentclass{article}

\PassOptionsToPackage{numbers, compress}{natbib}

    \usepackage[final]{neurips_2020}

\usepackage[utf8]{inputenc} 
\usepackage[T1]{fontenc}    
\usepackage{hyperref}       
\usepackage{url}            
\usepackage{booktabs}       
\usepackage{amsfonts}       
\usepackage{nicefrac}       
\usepackage{microtype}      
\usepackage{subcaption}
\usepackage{natbib}
\usepackage{graphicx}
\usepackage{xcolor}
\usepackage[ruled,vlined]{algorithm2e}
\usepackage{amsmath}
\usepackage{makecell}

\DeclareMathOperator*{\argmax}{arg\,max}

\title{Predicting Human Strategies in Simulated Search and Rescue Task} 
\author{%
  Vidhi Jain\\
  Carnegie Mellon University\\
  \texttt{vidhij@andrew.cmu.edu} \\
   \And
   Rohit Jena \\
  Carnegie Mellon University\\
  \texttt{rjena@andrew.cmu.edu} \\
  \And
  Huao Li  \\
  University of Pittsburgh\\
   \texttt{hul52@pitt.edu} \\
   \And
  Tejus Gupta\\
Carnegie Mellon University\\
   \texttt{tejusg@andrew.cmu.edu} \\
   \And
   Dana Hughes\\
Carnegie Mellon University\\
   \texttt{danahugh@andrew.cmu.edu} \\
       \And
   Michael Lewis\\
  University of Pittsburgh\\
   \texttt{ml@sis.pitt.edu} \\
        \And
   Katia Sycara \\
Carnegie Mellon University\\
   \texttt{katia@cs.cmu.edu} 
}

\begin{document}

\maketitle

\begin{abstract}

In a search and rescue scenario, rescuers may have different knowledge of the environment and strategies for exploration. Understanding what is inside a rescuer's mind will enable an observer agent to proactively assist them with critical information that can help them perform their task efficiently. To this end, we propose to build models of the rescuers based on their trajectory observations to predict their strategies. In our efforts to model the rescuer's mind, we begin with a simple simulated search and rescue task in Minecraft with human participants. We formulate neural sequence models to predict the triage strategy and the next location of the rescuer. As the neural networks are data-driven, we design a diverse set of artificial ``faux human'' agents for training, to test them with limited human rescuer trajectory data. To evaluate the agents, we compare it to an evidence accumulation method that explicitly incorporates all available background knowledge and provides an intended upper bound for the expected performance. Further, we perform experiments where the observer/predictor is human. We show results in terms of prediction accuracy of our computational approaches as compared with that of human observers. 

\end{abstract}

\section{Introduction}

%
Consider a scenario where an office building gets damaged during an earthquake and has  victims who need to be rescued. Victims would have varying degrees of injury, and therefore human rescuers have to develop appropriate search, navigation and victim triaging strategies so as to save the largest number of victims in limited time. The earthquake may have caused serious structural perturbations, such as blockages that limit the ability of the rescuers to reach certain areas.


 
{
In this paper, we provide initial results of a computational agent that observes the environment and the behavior of a human rescuer (in a simulation environment) and predicts future actions of the rescuer. This is the first step towards the development of an agent that can provide anticipatory assistance to the human. Anticipatory assistance is important especially in dynamic and dangerous situations. By  predicting future human actions, the agent can warn the human to avoid recently collapsed and unstable areas, or give advice to mitigate human cognitive limitations, such as limited memory of confusion over current localization, thus helping the human avoid duplicate  effort by revisiting areas that had already been searched. 
We evaluate the prediction accuracy of evidence accumulation approaches and neural sequence models on these tasks. Further, to ground the performance results of these methods, we compare their accuracy with human observers who saw the sequential actions taken by another human participant. }

Our specific scenario considers a 3D simulation of a realistic office building with several rooms and corridors where the disaster has taken place. 
For simplicity, victims are denoted as blocks in our environment. Some critically injured victims (denoted as yellow blocks) take more time to triage and may expire sooner than other victims (denoted as green blocks). Before starting the mission, we provide the rescuer with the building's original floorplan so that they can plan their route. During the mission, they may encounter several changes due to the damage, such as wall collapse or opening/holes in the wall.
We incentivize the rescuers to prioritize the critically injured victims by rewarding them more points for saving them.


Neural sequence models have been successful across diverse domains, and therefore we consider them for predicting strategies in our setting. Although such models learn implicitly from data, collecting large amount of training data from rescuers is laborious and expensive in our case.
So, we create diverse faux human agents to augment our human dataset to train different neural sequence methods to infer rescuers' navigation and triage strategies.
To evaluate our approach, we compare it to an evidence accumulation method where we explicitly incorporate all available background knowledge and thus provide an upper bound for the expected performance of the neural sequence approaches. 

Our main contributions are summarized as follows:


\begin{itemize}
    \item We develop AI models to predict the triage and navigation strategies of a human rescuer based on observed trajectories in a simulated search and rescue task, and compare the predictive ability of the AI models with those of human observers.
    \item We develop faux human trajectories to train neural sequence models, which are then evaluated on real human data. 
    \item We propose the evidence accumulation approach to incorporate all the available domain knowledge explicitly
    and thus, provide an upper bound on the expected performance of neural sequence models on the strategy prediction task.
\end{itemize}

\section{Related Work}


Disaster scenario in indoor environments has been considered and looked into by 
\cite{sohn2019deep}, \cite{dubey2019cognitive} where the crowd flow has been analyzed with simulated agents. 
Though these techniques assess simple fork in the corridor  structures for navigation with simulated agents only, we study human participants and their biases in a larger simulated setting as an office building with many such structures. 






Neural sequence models have shown promising results in several domains like trajectory prediction \cite{giuliari2020transformer} \cite{monti2020dagnet}. 
In particular, the transformer \cite{vaswani2017attention} is a popular attention mechanism for sequence modeling tasks and achieves state-of-the-art results on various benchmarks. But these models require a large amount of training data. As it is particularly challenging to obtain large human data in our setting, we discuss approaches with limited human data and our design of faux-human agents. 

LSTMs do not deal with irregularly sampled observations by default. To alleviate this problem, a model-agnostic representation of time \cite{time2vec} can be learnt and fed as an additional input.
 An alternative approach is to update the hidden state of the predictor between observations, such as through the use of an learned ordinary differential equation~\cite{rnnode}.

Methods based on Theory of Mind (ToM) framework reason in joint belief-intent space to reason about the demonstrator's behavior. Previous work \cite{baker2011bayesian} has shown that inferences using ToM models closely match predictions of human observers. However, these results were demonstrated in smaller settings, and it is challenging to scale Bayesian inference using ToM models to large environments such as ours. 
Incorporating ToM with neural networks is shown to be  successfully to reason about machine agents, where ground truth about the internals of the decision making are available. But these approaches are yet to be applied to reason about human mental state. 
Unlike settings with a single goal in 11x11 sized grid-worlds without any structural priors \cite{mtom2018corl}, we consider a more realistic multi-goal task setup where the agent can be attributed with different types of strategies. 


The challenge of understanding humans' beliefs from behavior has also been studied in \cite{sidreddy2018} where humans are assumed to act optimally but have incorrect knowledge of environment dynamics. We observe a similar case in our task setup where differing internal perspective on task complexity leads to variations across human behaviors.


\section{Approach}

To train the neural sequence models, we require a large amount of data. Since it is expensive to collect such a large dataset of human behavior in our setting, we augment our dataset of trajectories with a collection of faux-human agents. The faux-human agents are AI agents designed to quickly search the map and triage as many victims as possible. However, humans do not always act rationally, and therefore, these naive faux-human agents' behavior may be quite different from human behavior. We ameliorate this issue by collecting a small set of pilot human data and incorporating the rescuers' observed biases into the faux-human agents. 
Some of the observed biases in the rescuer's decision-making were: (1) choosing subgoals in a soft-optimal manner (modeled using Boltzmann distribution), (2) planning over room sequences instead of low-level actions, and (3) using greedy frontier-based search for short-horizon combined with long term planning aligned with the cliques in the graph representation of areas of a map. 
We thus obtain a rich and diverse set of faux-human agents that incorporate human biases while optimizing for the task objective. 

We model the rescuer's intent across the victim saving strategies, and navigation behavior in terms of next area to visit.
First, we consider two victim saving strategies - saving the critically injured first (selective) or saving whoever comes first (opportunistic)
These preferences tend to change in humans with time and proximity to the victims, making it a challenging sequential binary prediction task.
Second, given the area segments of the rooms and corridors
from the original floorplan, we formulate the next location prediction as a multi-class classification problem. 

We evaluate neural sequence models like RNNs and multi-head attention based transformer model \cite{vaswani2017attention} on our dataset. We compare this approach with a rule-based evidence accumulation method for prediction. This rule-based system is effective when we know the relevant evidence to track based on the rescuer's decision-making model and its knowledge state. In contrast, the neural sequence model learns what evidence to use and how to use it in an end-to-end manner using data.

\subsection{Evidence accumulation approach}
In this approach, we are explicitly incorporating full knowledge of SAR task and admissible strategies. We
maintain a belief over the likelihood over each of the classes for the strategy prediction per human trajectory.
 At initial state with no evidence, let the belief vector have uniform likelihood for each strategy as a prior. 
 We assume access to a library (or a look-up table) for evidences $e_i$ and their corresponding operation $f_i$ to update the belief. 
Throughout the rescuer's trajectory, this approach provides a way to update the likelihood sequentially depending on the evidence for each strategy/intent and predict the one which is the most likely, as shown in algorithm \ref{algorithm1}. 

\begin{algorithm}[H]
\SetAlgoLined
\KwResult{Most likely condition/strategy for every timestep in the trajectory}
 Rescuer's Trajectory $\tau_0^n$ where $n$ is the length of the trajectory\;
 Evidence Library $E = \{
 e_1 : f_1(\mathbf{b}),
 e_2 : f_2(\mathbf{b}),
 \cdots ,
 e_m : f_m(\mathbf{b}) 
 \}$ where $m$ are total evidences\;
 Condition set of strategy/intent $\mathcal{C} $ of size $d$\;
 Belief vector $\mathbf{b_t}=[1/d, 1/d, \cdots, 1/d]$ at $t=0$\;
  \For{timestep $t \in \{0, \cdots n\}$ }{
    \If{evidence $e_t$ found in $\tau_0^t$}{
      Obtain a function from the evidence library as $f_t =     E(e_t)$\;
      Find the belief vector as $\mathbf{b_{t+1}} = f_t(\mathbf{b_t})$\;
     }
   $i = \argmax \mathbf{b_{t+1}}$\;
  most likely condition at time $t = \mathcal{C}[i]$ \;
 }
 \caption{Evidence accumulation algorithm to predict the most likely strategy/intent}
 \label{algorithm1}
\end{algorithm}


Though the evidence accumulation algorithm depends on knowing every $e_i$ and its appropriate operation $f_i$ to update the belief, 
it provides an upper bound on the prediction, limited with only individual differences and momentary variations. For specific evidences used for triage and next location prediction, refer to the appendix \ref{sec:evidence}.



\subsection{Neural sequence models}
In settings where we have sufficient data in terms of the rescuer's trajectory but with no knowledge of specific evidences for prediction, we need to learn to infer them implicitly from their trajectory data. In this case, we train neural sequence models on faux human trajectories to evaluate their performance on the human data.


For victim triaging strategy, we train the three variants of recurrent neural networks, namely LSTM + RNNDecay \cite{rnnode}, LSTM + RNN-ODE \cite{rnnode} and LSTM + Time2Vec \cite{time2vec}. We observed the best performance of LSTM + Time2Vec \cite{time2vec} and use this model for comparison in table \ref{compare-table}. For detailed analysis of all three methods, refer table \ref{compare-neural-rnn-triage} in appendix. The input to each of the models is a sequence of observations, each of which consists of vector representation of the time, the 2D coordinates, and the condition of the victim seen (critically injured or not). 


For the next location prediction, 
we formulate the input as a sequence of areas visited. 
When the rescuer enters a new area segment, we take as input all the previously visited areas to predict the next area that they will transition to. 
We use a 2-layered 2-head Transformer model  \cite{vaswani2017attention}, which is neural network of encoder – decoder structure with partial masking of input for sequential prediction. 
Unlike the evidence accumulation approach, we do not provide the map connectivity. Rather, it is inferred from the input sequences by the network. 
We provide further details for both the models in appendix \ref{sec:neural}.

\section{Experiment setup}
\subsection{Task scenario}
\begin{figure}
\centering
\begin{subfigure}{.45\textwidth}
\centering
\includegraphics[width=6cm]{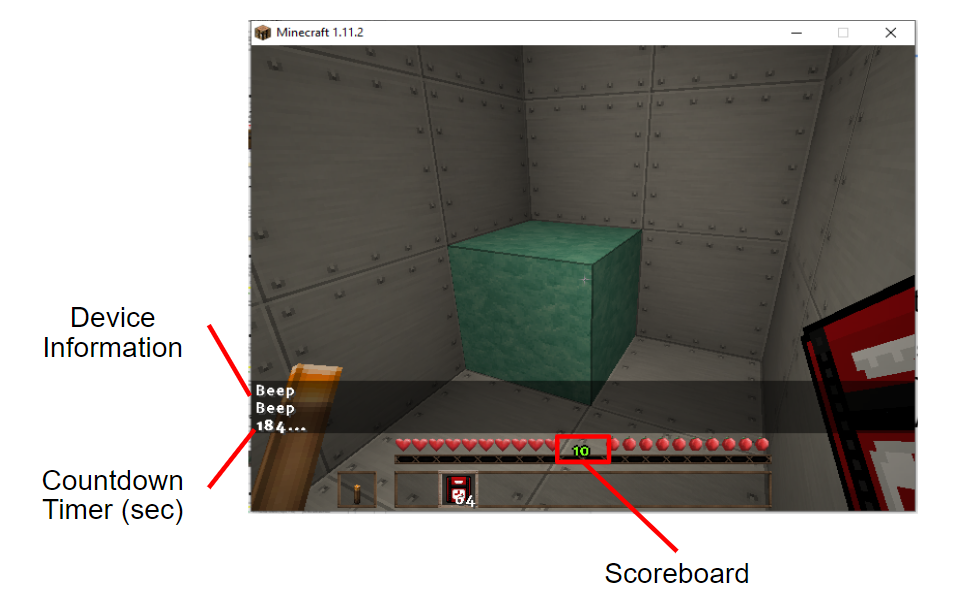}
\label{player_view}
\caption{3D Minecraft view to the human participants}
\end{subfigure}
\begin{subfigure}{.45\textwidth}
\centering
\includegraphics[width=6cm]{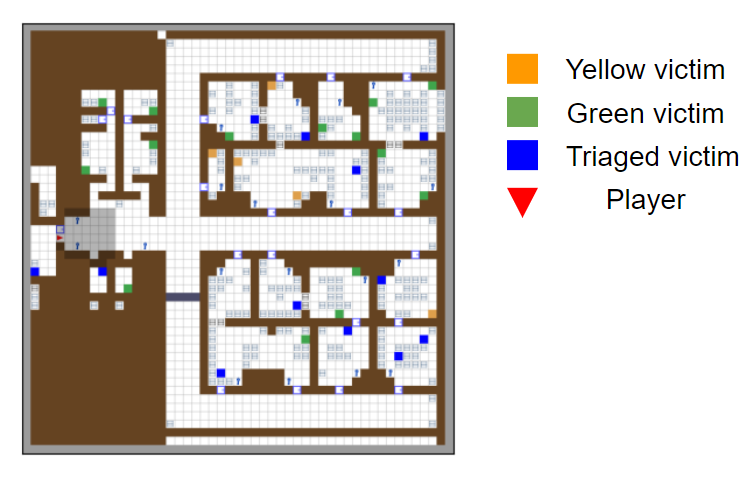}
\label{map_view}
\caption{2D map layout of the environment}
\end{subfigure}
\caption{
Human participants as rescuer see the view in (a). Human observers view the replay of the trajectory as in (a) and the bird's eye view of the environment as in (b).
}
\label{exp_fig}
\end{figure}

We build a damaged office building in Minecraft environment as the task scenario for rescuers. The game screen and a 2D map layout of this environment are shown in Fig  \ref{exp_fig}. 
The scenario in the Minecraft environment represents a structurally damaged office building after an unspecified incident. Initially, it contains 26 area segments consisting of corridors, rooms, and elevators. The current building layout and segment connectivity were changed by perturbations such as collapses, wall openings, and sporadic fires. There are 20 injured victims inside the building who need to be rescued. Out of these, five victims are severely injured (yellow) and might die if not treated in time. Other victims are denoted in green.
Both victims depend on the first responders’ help to stabilize and evacuate out of the building.



\subsection{Human trajectories collection}

Eight participants were recruited to collect real human trajectories in the search and rescue task. Participants were given 
detailed instructions on the search task and the virtual environment they would interact with. Each participant completed the same searching task 3 times on maps with 3 different perturbations. In each of the 15-min trial, human participants were asked to control a rescuer avatar and search for victims in the building. Player's position, interaction history and field of view were recorded as the game log in a sampling rate of 5Hz. In total 24 trajectories were collected from human players, 6 out of which were excluded from following analysis because of incomplete data record.


\subsection{Human observer experiment}
To provide a sound baseline of our agent prediction, we provide the same Theory of Mind inference task to human observers. The gameplay screen recordings and minimap videos were segmented by `decision points' at which behaviors occur such as spotting a victim or leaving a room. Segments were viewed in the order recorded so that prior segments can inform judgments. Human observers were asked to supply a prediction of the expected action and choose among alternative beliefs and intentions from menus. The action taken was then presented at the start of the following sequence. 50 workers on Amazon MTurk participated in the observation experiment. They were asked to predict five different types of rescuers' action in corresponding decision points. Refer to appendix \ref{sec:human-exp} for a detailed definition of decision points and online survey design.

\vspace*{-10pt}
\section{Results}

The results shown in Table \ref{compare-table} highlight the prediction performance of our computational agent as compared to the human observers. 
To evaluate our approach, we compare it to an evidence accumulation method that explicitly incorporates all available background knowledge and thus provides an upper bound for the expected performance. 
The human observers were evaluated on a selected set of 10 decision timesteps for 18 human trajectories, while the two approaches were tested at every decision timestep relevant for triaging and next location prediction. 

For triage strategy prediction, we obtain a high performance of $98.80\%$ with evidence accumulation approach.  This is a sequential binary classification task for which the selected evidence and belief update functions incorporate most of the domain knowledge. The neural sequence modeling approach implicitly infers from the trajectory observations and performs with an accuracy of 70.74\%. 

The next location prediction is a 
multi-class classification, and a harder task than triage prediction. In this case, we found that devising exact evidences is challenging as human rescuers tend to be stochastic in the next location selection and therefore, we observe about $68.53\%$ average prediction accuracy. The neural sequence models achieves a close performance with accuracy of $66.98\%$. 

Overall, we observe that the computational methods outperform humans at both the prediction tasks.
The results highlight evidence accumulation is a good enough approach to incorporate domain knowledge in limited data setting and can serve as an upper bound to the neural sequence model's performance. 
Further, neural sequence models can perform comparable to evidence accumulation approaches when the tasks get more complex, which makes it challenging to encode all the required evidences and their corresponding belief update functions for strategy prediction.

\begin{table}
  \caption{Comparison of strategy prediction accuracy. 
  }
  \label{compare-table}
  \centering
  \begin{tabular}{lrr}
    \toprule
    Prediction Method    &  Triage strategy & Next location \\
    \midrule
    Human observers          &  65.50\% & 58.20\%      \\
    Evidence accumulation    & {98.80}\% & 68.53\%       \\
    Neural sequence        & {70.74}\%   & 66.98\%       \\
    \bottomrule
  \end{tabular}
\end{table}
\section{Discussion}
Our work aims to develop artificial agents that provide anticipatory assistance to human rescuers in disaster relief.  Anticipatory assistance is important especially for dangerous and dynamic environments, such as disaster relief. To provide such assistance,  the agent would observe the human and the environment and try to predict human's actions from observed behavior, so as to be able to e.g. warn the human of dangerous dynamic changes or mitigate shortcoming resulting from human cognitive limitations, such as limitations of memory or confusion over localization (e.g. the human spends redundant effort in revisiting places they have already searched). We have obtained initial encouraging results compared with humans in the role of the assisting agent. We  aim to refine the computational  models so that better results would be obtained. Additionally, we will study agent assistance for human teams of rescuers. We expect this to be a more challenging task that would provide opportunities for new insights and development of novel computational neural network architectures.



\section*{Broader Impact}

The proposed work, though in its initial stages promises to contribute to science and technology for decision support in search and rescue and disaster relief, thus helping to save human lives. We do not foresee any negative impacts or outcomes of this research.
As we observed that both the AI models perform better than human observers, this provides evidence that some tasks related to predicting rescuer behavior may be offloaded to AI agents, thus reducing the cognitive load of human observers performing potentially several tasks in parallel.




\begin{ack}
This material is based upon work supported by the Defense Advanced Research Projects Agency (DARPA) under Contract No. HR001120C0036. Any opinions, findings and conclusions or recommendations expressed in this material are those of the author(s) and do not necessarily reflect the views of the Defense Advanced Research Projects Agency (DARPA).



\end{ack}



\bibliographystyle{plainnat}
\bibliography{sample}

\begin{thebibliography}{10}
\providecommand{\natexlab}[1]{#1}
\providecommand{\url}[1]{\texttt{#1}}
\expandafter\ifx\csname urlstyle\endcsname\relax
  \providecommand{\doi}[1]{doi: #1}\else
  \providecommand{\doi}{doi: \begingroup \urlstyle{rm}\Url}\fi

\bibitem[Baker et~al.(2011)Baker, Saxe, and Tenenbaum]{baker2011bayesian}
Chris Baker, Rebecca Saxe, and Joshua Tenenbaum.
\newblock Bayesian theory of mind: Modeling joint belief-desire attribution.
\newblock In \emph{Proceedings of the annual meeting of the cognitive science
  society}, volume~33, 2011.

\bibitem[Dubey et~al.(2019)Dubey, Sohn, Hoelscher, and
  Kapadia]{dubey2019cognitive}
Rohit~K. Dubey, Samuel~S. Sohn, Christoph Hoelscher, and Mubbasir Kapadia.
\newblock Cognitive agent based simulation model for improving disaster
  response procedures, 2019.
\newblock URL \url{https://arxiv.org/abs/1910.00767}.

\bibitem[Giuliari et~al.(2020)Giuliari, Hasan, Cristani, and
  Galasso]{giuliari2020transformer}
Francesco Giuliari, Irtiza Hasan, Marco Cristani, and Fabio Galasso.
\newblock Transformer networks for trajectory forecasting, 2020.
\newblock URL \url{https://arxiv.org/abs/2003.08111}.

\bibitem[Kazemi et~al.(2019)Kazemi, Goel, Eghbali, Ramanan, Sahota, Thakur, Wu,
  Smyth, Poupart, and Brubaker]{time2vec}
Seyed~Mehran Kazemi, Rishab Goel, Sepehr Eghbali, Janahan Ramanan, Jaspreet
  Sahota, Sanjay Thakur, Stella Wu, Cathal Smyth, Pascal Poupart, and Marcus
  Brubaker.
\newblock Time2vec: Learning a vector representation of time.
\newblock \emph{arXiv preprint arXiv:1907.05321}, 2019.

\bibitem[Monti et~al.(2020)Monti, Bertugli, Calderara, and
  Cucchiara]{monti2020dagnet}
Alessio Monti, Alessia Bertugli, Simone Calderara, and Rita Cucchiara.
\newblock Dag-net: Double attentive graph neural network for trajectory
  forecasting, 2020.
\newblock URL \url{https://arxiv.org/abs/2005.12661}.

\bibitem[Rabinowitz et~al.(2018)Rabinowitz, Perbet, Song, Zhang, Eslami, and
  Botvinick]{mtom2018corl}
Neil~C. Rabinowitz, Frank Perbet, H.~Francis Song, Chiyuan Zhang, S.~M.~Ali
  Eslami, and Matthew Botvinick.
\newblock Machine theory of mind.
\newblock \emph{CoRR}, abs/1802.07740, 2018.
\newblock URL \url{http://arxiv.org/abs/1802.07740}.

\bibitem[Reddy et~al.(2018)Reddy, Dragan, and Levine]{sidreddy2018}
Siddharth Reddy, Anca~D. Dragan, and Sergey Levine.
\newblock Where do you think you're going?: Inferring beliefs about dynamics
  from behavior.
\newblock \emph{CoRR}, abs/1805.08010, 2018.
\newblock URL \url{http://arxiv.org/abs/1805.08010}.

\bibitem[Rubanova et~al.(2019)Rubanova, Chen, and Duvenaud]{rnnode}
Yulia Rubanova, Ricky~TQ Chen, and David Duvenaud.
\newblock Latent odes for irregularly-sampled time series.
\newblock \emph{arXiv preprint arXiv:1907.03907}, 2019.

\bibitem[Sohn et~al.(2019)Sohn, Moon, Zhou, Yoon, Pavlovic, and
  Kapadia]{sohn2019deep}
Samuel~S. Sohn, Seonghyeon Moon, Honglu Zhou, Sejong Yoon, Vladimir Pavlovic,
  and Mubbasir Kapadia.
\newblock Deep crowd-flow prediction in built environments, 2019.
\newblock URL \url{https://arxiv.org/abs/1910.05810}.

\bibitem[Vaswani et~al.(2017)Vaswani, Shazeer, Parmar, Uszkoreit, Jones, Gomez,
  Kaiser, and Polosukhin]{vaswani2017attention}
Ashish Vaswani, Noam Shazeer, Niki Parmar, Jakob Uszkoreit, Llion Jones,
  Aidan~N Gomez, {\L}ukasz Kaiser, and Illia Polosukhin.
\newblock Attention is all you need.
\newblock In \emph{Advances in neural information processing systems}, pages
  5998--6008, 2017.

\end{thebibliography}

\section*{Appendix}
\subsection{Evidence accumulation approach}
\label{sec:evidence}
\textbf{For victim triage prediction,} an event when the rescuer ignores less-critical victim is an evidence, say $e_1$.
The corresponding update $f_1$ will increase the likelihood of the selective triage strategy in the belief vector $\mathbf{b_t}$ for the current time $t$.
Similarly, if a less-critical victim is triaged upon sight, it is evidence $e_2$ for increasing the likelihood of opportunistic victim triage strategy.



\textbf{For next location prediction, }
the map connectivity of the perturbed environment serves as a useful domain knowledge.
We capture the layout as a graph where nodes are the area segments and the edges denote that areas are connected.  At each step $e_i = (x, y)$ where $x,y$ are position coordinates of the rescuer, which is in turn used to infer the area segment. The corresponding $f_i$  increases the likelihood of each connected area 
depending on (1) the out degree of the node representing the area, and (2) the distance of the rescuer's position to that area.
Another evidence is based on the fact that the rescuer is on an exploration task, which means they are unlikely to go to already visited areas and so, we decrease the likelihood of the area segments that have been visited.

\subsection{Architecture details for neural-based sequence models}
\label{sec:neural}
\textbf{Victim triage strategy: }We formulate the input to the neural network model as a sequence based on every event when a victim comes into the field of view of the rescuer, and when the rescuer finishes triaging a victim. 
We train LSTM + RNNDecay \cite{rnnode}, LSTM + RNN-ODE \cite{rnnode} and LSTM + Time2Vec \cite{time2vec}. 
For the most part, we use the original architecture's hyperparameters except choosing batch size 1, and decay half-life as 60. The networks are trained to minimize the cross entropy loss with the groundtruth label with Adam optimizer with AMSGrad technique with learning rate 3e-4 and weight decay for regularization as 1e-5.
\begin{table}[h]
  \caption{Comparison of baselines for triage strategy prediction}
  \label{compare-neural-rnn-triage}
  \centering
  \begin{tabular}{lrr}
    \toprule
    Prediction Method   &  Accuracy  \\
    \midrule
    Evidence accumulation    &  98.80\%\\
    LSTM + RNNDecay \cite{rnnode}   & {63.03}\% \\
    LSTM + RNN-ODE \cite{rnnode}   & {67.44}\% \\
    LSTM + Time2Vec \cite{time2vec}    & {70.74}\% \\
    \bottomrule
  \end{tabular}
\end{table}

\textbf{Next room prediction: } We formulate the input as a sequence of rooms that the player moves to. For example such a sequence from a rescuer's trajectory is: `Stairwell Starting Point', `Center Hallway Lobby', ``Women's Bathroom", `Center Hallway Lobby', `Elevator 1', `Center Hallway Lobby', `Center Hallway', `Room 211', `Room 213', `Room 218', `Right Hallway', `Room 216', `Room 209', `Center Hallway', `Room 208', `Center Hallway', `Room 210', `Room 207', `Room 210', `Center Hallway', `Room 215', `Center Hallway', `Room 208', `Room 203', `Room 201', `Left Hallway', `Center Hallway Lobby', ``Men's Bathroom", `Center Hallway Lobby', `Left Hallway', `Room 205', `Left Hallway', `Left Hallway', `Center Hallway', `Room 211', `Room 213', `Room 218', `Right Hallway', `Room 220' .

Since there are 26 area segments in the chosen map, we learn to predict the next room given a sequence of previously visited rooms. 
We use a neural network of 2-head, 2-layer transformer model, that is an encoder – decoder structure with multi-head attention over the masked sequence. We learn the first layer as a 26 dimensional embedding of each room area in the masked input sequence. This is processed by Transformer encoder with hidden state dimension as 8. We use 5 steps for propagation through time. 
The rest of the hyperparameter are the same as existing pytorch code on transformer models \footnote{ \url{https://github.com/pytorch/examples/tree/master/word_language_model}}. 
Finally, the output is decoded by a linear weights such that each dim corresponds to the log likelihood of that being the next room.


\subsection{Human observation experiments} \label{sec:human-exp}
\subsubsection{Materials}
In total 18 rescuer trajectories were used in the human observation experiment. Depending on the performance of original rescuer, the length of each trajectory range from 8 minutes to 15 minutes. Based on the collected human trajectories, we generated following materials: game screen recording videos, dynamic minimap videos and a static building layout image. Human observers can watch the first person screening recording of rescuers to understand what they were doing, and refer to the dynamic/static maps to help locate the rescuers' current location and navigation path. Video materials were were segmented by `decision points' at which behaviors occur such as spotting a victim or leaving a room. The decision points are explained below. 

\begin{itemize}
\item Triage decision points
    \begin{itemize}
        \item Definition: When a victim block enters rescuer's FOV.
        \item Prediction task: Will the rescuer triage the victim in front of him?
    \end{itemize}
\item Navigation decision points
    \begin{itemize}
        \item Definition: When a room entrance (door/hole) enters rescuer's FOV.
        \item Prediction task: Will the rescuer enter the room in front of him?
    \end{itemize}
\item General decision points
    \begin{itemize}
        \item Definition: When the rescuer finishes triaging a victim / leaving a room.
        \item Prediction task: 
        \\What is the next location the rescuer will go?
        \\What is the rescuer's triage strategy? 
        \\What is the rescuer's knowledge condition?
    \end{itemize}
\end{itemize}
At each of the different decision points, human observers were given different prediction tasks including predicting next room and triage strategy of the securer etc. They were asked choose among alternative locations or strategies from menus. Video segments were viewed in the order recorded so that prior segments can inform judgments. The actual action taken by the rescuer in video was then presented at the start of the following sequence. The total number of decision points in one trajectory is around 300, which is too demanding for human observers to annotate. We sampled 10 decision points for each type and generated 30 video segments with corresponding prediction questions for each trajectory.

\subsubsection{Procedure}
50 human observers were recruited from Amazon Mechanical Turk. Participant accessed the online survey on their own computer. A detailed instruction was given to observers about the searching environment and the prediction task they need to complete. Then the observers need to pass a quiz about basic knowledge of our experiment in order to process to the formal task. Each observer was assigned one trajectory from human rescuer. In each of the 30 trials, human observers were presented a video clip and the prediction questions. The length of this human observation experiment is around 45 minutes.

\end{document}